\documentclass{article}

\usepackage{microtype}
\usepackage{graphicx}
\usepackage{subfigure}
\usepackage{booktabs} 

\usepackage{hyperref}

\usepackage[accepted]{icml2024}

\usepackage{amsmath}
\usepackage{amssymb}
\usepackage{mathtools}
\usepackage{amsthm}
\usepackage[T1]{fontenc}
\usepackage[subtle]{savetrees}
\usepackage[capitalize,noabbrev]{cleveref}

\theoremstyle{plain}
\newtheorem{theorem}{Theorem}[section]

\theoremstyle{definition}
\newtheorem{definition}[theorem]{Definition}
\newtheorem{assumption}[theorem]{Assumption}
\theoremstyle{remark}

\usepackage[textsize=tiny]{todonotes}

\icmltitlerunning{Exact Conversion of In-Context Learning to Model Weights}
\begin{document}

\twocolumn[
\icmltitle{Exact Conversion of In-Context Learning to Model Weights in Linearized-Attention Transformers}



\icmlsetsymbol{equal}{*}

\begin{icmlauthorlist}
\icmlauthor{Brian K Chen}{nus,sch}
\icmlauthor{Tianyang Hu}{huawei}
\icmlauthor{Hui Jin}{huawei}
\icmlauthor{Hwee Kuan Lee}{nus,sch,ntu,seri,sirlai,sics}
\icmlauthor{Kenji Kawaguchi}{nus}
\end{icmlauthorlist}

\icmlaffiliation{huawei}{Huawei Noah's Ark Lab}
\icmlaffiliation{nus}{National University of Singapore}
\icmlaffiliation{sch}{Bioinformatics Institute, Agency for Science, Technology and Research (A*STAR)}
\icmlaffiliation{ntu}{Nanyang Technological University}
\icmlaffiliation{seri}{Singapore Eye Research Institute}
\icmlaffiliation{sirlai}{Singapore International Research Laboratory on Artificial Intelligence}
\icmlaffiliation{sics}{Singapore Institute for Clinical Sciences}

\icmlcorrespondingauthor{Tianyang Hu}{hutianyang.up@outlook.com}
\icmlcorrespondingauthor{Kenji Kawaguchi}{kenji@comp.nus.edu.sg}

\icmlkeywords{Machine Learning, ICML}

\vskip 0.3in
]



\printAffiliationsAndNotice{}  

\begin{abstract}

In-Context Learning (ICL) has been a powerful emergent property of large language models that has attracted increasing attention in recent years.  In contrast to regular gradient-based learning, ICL is highly interpretable and does not require parameter updates.  In this paper, we show that, for linearized transformer networks, ICL can be made explicit and permanent through the inclusion of bias terms.  We mathematically demonstrate the equivalence between a model with ICL demonstration prompts and the same model with the additional bias terms. Our algorithm (ICLCA) allows for exact conversion in an inexpensive manner. Existing methods are not exact and require expensive parameter updates. We demonstrate the efficacy of our approach through experiments that show the exact incorporation of ICL tokens into a linear transformer. We further suggest how our method can be adapted to achieve cheap approximate conversion of ICL tokens, even in regular transformer networks that are not linearized.  Our experiments on GPT-2 show that, even though the conversion is only approximate, the model still gains valuable context from the included bias terms.
\end{abstract}

\section{Introduction}
Large Language Models (LLMs) have displayed incredible performance and capabilities with the introduction of models such as BERT \citep{devlin2018bert} and GPT \citep{radford2018improving, brown2020language}. One such ability is In-Context Learning (ICL), which involves learning to generalize and make predictions on new tasks based on demonstration examples provided in the prompt.  Numerous studies have shown LLMs solving increasingly complex tasks with ICL as compared to fine-tuning using labeled data \citep{wei2022chain}.

In-context learning is an attractive proposition compared to traditional parameter-based model update methods such as fine-tuning \citep{wei2021finetuned, ouyang2022training} or knowledge editing \citep{dai2021knowledge, meng2022locating, meng2022mass}.  
Demonstration examples can be directly included in the prompt. There is no need to edit the parameters of the model which can be costly when considering models with billions of parameters such as OpenAI's GPT-4 and Google's Bard.  
In terms of performance, ICL prompting methods \citep{wei2022chain, zhou2022least} also demonstrate stronger generalization and more precise control of the model behavior \citep{zheng2023can}.  Such an advantage is expected to be more prominent as LLMs get larger and more powerful. 
In addition, ICL presents a degree of interpretability when incorporating demonstration examples into the model.  The examples often take the form of natural language prompts or other readily understandable inputs that can be understood by humans. This is often not the case when working with machine learning.

Despite its promising qualities, there are clear shortcomings to the ICL paradigm compared to the traditional gradient-based fine-tuning approaches.  While not requiring costly parameter updates like other methods, ICL does not permanently modify the model weights.  All included contextual information is only temporary.  Demonstration examples are only incorporated on a case-by-case basis and will not be applied to future input prompts.
Studies on permanently including ICL information in the model weights are still limited. There are some attempts aiming to distill the context information from ICL prompts into the model parameters \citep{askell2021general, choi2022prompt, snell2022learning}. However, these so-called context distillation methods are largely empirical and lack theoretical guarantees. 

In this study, we attempt to incorporate ICL demonstration prompts into the parameters of an LLM through the addition of a bias term to the attention module.  We do so for a specific class of transformers with linearized attention modules which have already shown promise in real-world applications for improving the efficiency of LLMs. 

There are a few key advantages of the ICL conversion method to be introduced in Section \ref{sec: extra param for ext}. 
First, unlike the fine-tuning data, which is often just another sampled dataset, the ICL prompts used here are highly interpretable and often meaningful in the context of natural languages.
Since we are doing exact conversion rather than fine-tuning, we will not face issues with algorithmic convergence or stability that arise with fine-tuning methods.  As long as the ICL ability of the original model is strong and we have suitable ICL prompts, our method will be effective.
Second, our proposed method is simple to implement. All that is required is a linear update to the biases of the network, which is computationally very inexpensive compared to conducting gradient descent methods. 
The converted extra bias term only amounts to about 1\% of the total number of parameters, which is very efficient to store.
Furthermore, such extra weights can be combined freely to incorporate more in-context prompts or thrown away to recover the original model. 
Consequently, ICL conversion methods shape up to be a powerful new tool for incorporating new information into existing LLMs in a cheap, interpretable fashion.

In addition to an exact conversion method for transformers with linearized attention modules, we also provide an approximate conversion method for regular transformers in Section \ref{sec: approx}.  Transformers with softmax attention remain the most popular architecture for most state-of-the-art models so even an approximate conversion method bears significance.  While we can only achieve approximate conversion in this case, we show experimentally that the model retains some degree of In-context information by including our bias terms.  Approximate conversion of regular transformers is also easy to implement and readily interpretable.

Our work presents the following key contributions:

\begin{itemize}
    \item We demonstrate how ICL relationships are captured between the Key and Value matrices of the attention module.
    
   \item We present an exact conversion method that captures the ICL prompts into the model weights by introducing a novel bias term into the model architecture.  This conversion is done for models based on linearized attention.  This methodology is theoretically and mathematically supported in contrast to existing methods.
   
   \item We extend our methodology to an approximate conversion method that covers a wider setting. This includes the regular transformer with softmax-based attention.
   
   \item We provide experimental justification for this methodology: 
   \begin{itemize}
       \item We showcase the exact conversion of the ICL prompt into a model with linear attention.
       \item We demonstrate an effective approximate conversion of ICL prompts within a pretrained GPT-2 model with regular attention.
   \end{itemize}
\end{itemize}

\section{Related Works}
\textbf{Transformers and Large Language Models} The transformer architecture was first introduced by Vaswani et al. \citeyearpar{vaswani2017attention} and has revolutionized multiple fields including, but not limited to, natural language processing, computer vision, time-series analysis, and bioinformatics. The field of natural language processing has been especially impacted by transformers being adopted in LLMs such as GPT \citep{radford2018improving} and BERT \citep{devlin2018bert}.  Through the adoption of the attention mechanism, transformer architectures are capable of capturing long-range dependencies of the input sequence and more sophisticated abilities such as reasoning \citep{bubeck2023sparks, abbe2023generalization, dziri2024faith, wang2024understanding, giannou2023looped, gao2024expressive}.  This was difficult for previous alternatives such as LSTMs \citep{LSTM} and RNNs \citep{rumelhart1985learning} due to their sequential nature.

\paragraph{Linear and Kernel Attention}
Within the transformer layer, the softmax function is usually identified as a bottleneck.  It introduces quadratic complexity in computation concerning input length.  As an alternative, studies have approximated the attention mechanism by the linear product \citep{irie2022dual, bahdanau2016neural, luong2015effective}. This is further extended to approximate the softmax by a kernel product
\citep{blanc2018adaptive, katharopoulos2020transformers}.  As a field growing in popularity, there have been multiple studies attempting to find the most appropriate kernel, either for training models from scratch \citep{chowdhury2021learning}, or for approximating the softmax. \citep{choromanski2020rethinking}.  In particular, RetNet \citep{sun2023retentive} claims to match state-of-the-art transformer architectures using just linear attention with some added tricks such as multi-scale retention. Currently, while linearized transformer networks still occupy the minority, the prevalence of lengthy prompts has shown how important it is to rescue the computational cost of LLMs.  We predict that as time passes, linearized transformer networks will grow a great deal both in proficiency and significance.

\textbf{In Context Learning and Context Distillation}
The ICL capability of LLMs has gained much traction since being demonstrated on GPT-3 \citep{brown2020language}.  It has been considered an emergent capability that only arises in models with many layers and an extremely large number of parameters \citep{dong2022survey}.  However, studies have demonstrated that ICL can occur in smaller models, albeit in a less sophisticated manner \citep{wei2023larger}.
Due to the rising interest in ICL, multiple works have explored why ICL works and how it can be improved. Among others, Garg et al. \citeyearpar{garg2022can}, Akyurek et al. \citeyearpar{akyurek2022learning} and Xie et al. \citeyearpar{xie2021explanation} constructed a Bayesian framework to explain how ICL works when the training and inference distributions are distinct. Bai et al. \citeyearpar{bai2023transformers} provided a more concrete mechanism and demonstrated how transformers can implement a large range of learning algorithms in context.  They also provided experimental results showing the strong in-context algorithm selection capabilities of standard transformer architectures.
To make the effect of ICL more permanent, various methods have been developed to distill in-context prompts into the models. 
Askell et al. \citeyearpar{askell2021general}, Choi et al. \citeyearpar{choi2022prompt} and Snell et al. \citeyearpar{snell2022learning} proposed various fine-tuning schemes to match the input-output relationships with ICL. This line of work is largely empirical and usually involves multiple LLMs, which are not efficient and lack theoretical guarantees.

Based on our observation, current studies attempt to encode ICL permanently through a form of fine-tuning with gradient descent.  Such approaches only amount to approximations and can be computationally very expensive on increasingly large models.  Our work seeks to take advantage of the structure of linear and kernel attention to achieve the exact conversion of ICL to the weights of our LLM thereby avoiding both limitations of existing methodologies.

\section{Preliminaries}
 Introduced in the seminal paper ``Attention is All You Need" \citep{vaswani2017attention}, transformer models handle sequence-to-sequence tasks more proficiently by leveraging the attention mechanism to weigh the influence of tokens against each other.  Traditional sequence-to-sequence models, such as GRUs and LSTMs, suffer from difficulties in capturing long-range dependencies due to the inherently sequential nature of their operations. Such a problem does not occur in Transformers because each token has access to all the other tokens along the sequence allowing distant relationships to be captured.

In this paper, we assume that the input of the transformer model is initially passed through an encoder layer to obtain a sequence of vector tokens $X = [X_{1},...,X_{N}]^{T}$ which are then fed into the transformer layers.  Each token has dimensionality $d_{in}$ such that $X \in \mathbb{R}^{N \times d_{in}}$.

\subsection{The Attention Mechanism} \label{sec:attn}
The core aspect of the transformer is the attention mechanism.  Inspired by the cognitive process of information retrieval in database systems, the attention mechanism uses query, key, and value matrices to emulate how we would sort information in real life within a large database.

While in practical applications we often conduct multi-headed attention, this study emphasizes single-headed attention due to its simplicity.  All following definitions and methods can be translated to multi-headed attention by addressing each attention head separately.
\begin{definition}[Single-headed Attention Layer]
    Given input data $X \in \mathbb{R}^{N \times d_{in}}$, a single attention layer is characterized by trainable matrices $W_{Q},W_{K}\in \mathbb{R}^{d_{in} \times d_{K}}, W_{V}\in \mathbb{R}^{d_{in} \times d_{V}}$.  The Single-headed attention layer takes the form:
\begin{equation}
\label{eqn:regular attention}
    O(Q,K,V) = \mathrm{Attn}(Q,K,V)=\mathrm{softmax}\left(\frac{QK^{T}}{\sqrt{d_{K}}}\right)V,
\end{equation}
where $O$ is the output, $Q = XW_{Q}$, $K = XW_{K}$ and $V = XW_{V}$.
\end{definition}

For any given matrix $Z$, let $Z_{i}$ return the $i$-th row as a vector and denote $sim(\cdot,\cdot)$ as a similarity score (Note vectors are all vertical). Drawing upon the work of Katharopoulos et al. \citeyearpar{katharopoulos2020transformers}, the single-headed attention layer can be generalized as follows:
\begin{equation}
\label{eqn:reg_attn}
    O^{T}_{i} = \frac{\sum_{j=1}^{N}\mathrm{sim}(Q_{i},K_{j})V^{T}_{j}}{\sum_{j=1}^{N}\mathrm{sim}(Q_{i},K_{j})}.
\end{equation}
This is the same as regular single-headed attention when $\mathrm{sim}(q,k) = \exp(\frac{q k^{T}}{\sqrt{d_{K}}})$.  Such an approach can be extended to a linearized (kernel) attention setting by approximating the similarity score $\mathrm{sim}(\cdot,\cdot)$ by a kernel $\mathrm{Ker}(\cdot,\cdot)$ with associated feature map $\phi(\cdot)$.  There are many different choices for $\phi(\cdot)$. For example, Katharopoulos et al. \citeyearpar{katharopoulos2020transformers} uses a simple $\phi(x) = \mathrm{elu}(x) +1$.  Other feature maps are also further studied showcasing varying levels of success \citep{zhang2024hedgehog,katharopoulos2020transformers,choromanski2020rethinking}.

This is further generalized into the following linearized attention module:

\begin{definition}[linearized attention Layer]
    Given input data $X \in \mathbb{R}^{N \times d_{in}}$, a single attention layer is characterized by trainable matrices $W_{Q},W_{K}\in \mathbb{R}^{d_{in} \times d_{K}}, W_{V}\in \mathbb{R}^{d_{in} \times d_{V}}$. The linearized attention layer takes the form of:
\begin{equation}
\label{eqn:lin_attn}
\begin{split}
O^{T}_{i} &= \frac{ \sum_{j=1}^{N} \phi(Q_{i})^{T}\phi(K_{j})V^{T}_{j}}{D_{1}(Q_{i})^{T}D_{2}(X)} \\
&=\frac{\phi(Q_{i})^{T}\sum_{j=1}^{N}\phi(K_{j})V^{T}_{j}}{D_{1}(Q_{i})^{T}D_{2}(X)},
\end{split}
\end{equation}
\end{definition}
where $\phi(\cdot)$ is the associated feature map and $D_{1}(Q_{i})^{T}D_{2}(X)$ is a normalizing term depending on the network architecture and input $X$.  Typically $D_{2}$ is a function of $K = XW_{K}$, but we keep it as a function of $X$ itself to capture a more general class of models. For instance, when implementing kernel softmax we would take $D_{1}(Q_{i}) = \phi(Q_{i})$, $D_{2}(X) = \sum_{j=1}^{N} \phi(K_{j}) = \sum_{j=1}^{N} \phi(X_{j}W_{K})$ and when implementing linear attention we would take $D_{1}(Q_{i})=1, D_{2}(X) = 1$. 

For simplicity, Katharopoulos et al. \citeyearpar{katharopoulos2020transformers} state the numerator is easier to understand when stated in the vectorized form:
\begin{equation}
\label{vectorizedattention}
    (\phi(Q)\phi(K)^{T})V = \phi(Q)(\phi(K)^{T})V.
\end{equation}

In real-world applications involving sequential data such as NLP, transformers are typically autoregressive.  In the regular attention setting this requires applying a mask to the attention score matrix $QK^{T}$ within the softmax function.  However, the linearized attention layer requires the following modification:
\begin{definition}[Masked linearized attention Layer]
\label{dfn:masked_lin_attn}
    Given input data $X \in \mathbb{R}^{N \times d_{in}}$ a single attention layer is characterized by trainable matrices $W_{Q},W_{K}\in \mathbb{R}^{d_{in} \times d_{K}}, W_{V}\in \mathbb{R}^{d_{in} \times d_{V}}$. The Masked linearized attention Layer takes the form of:

\begin{equation}
O^{T}_{i} = \frac{1}{D_{1}(Q_{i})^{T}D_{2}(X)}\phi(Q_{i})^{T}\sum_{j=1}^{i}\phi(K_{j})V^{T}_{j}.
\end{equation}    
\end{definition}

In this study, we introduce the concept of the Key-Value matrix.  The Key-Value matrix is not widely explored within existing literature largely because it does not exist within the classic attention mechanism. However, it is present in linearized attention models and is the portion of the model that captures the ICL information as we find in \S \ref{sec: extra param for ext}
\begin{definition}[Key-Value Matrix] \label{KV-matrix}
    Given the definition of linearized attention found in Equation \eqref{eqn:lin_attn}, we define the Key-Value matrix $A$ as:
   \setlength{\abovedisplayskip}{3pt}
    \begin{equation}
        A = \sum_{j=1}^{N} \phi(K_{j})V^{T}_{j}.
    \end{equation}
    \setlength{\belowdisplayskip}{3pt}
    The definition is the same for masked linearized attention except the sum is up to token $i$ rather than over the whole sequence with length $N$.
\end{definition}
We can see that the addition of extra tokens to the input will not change the $\phi(Q_{i})$ term but only the Key-Value matrix. Hence, conversion within the Key-Value matrix $A$ allows for conversion within the linearized attention Layer.

\subsection{In Context Learning}
In traditional supervised learning, models are generally trained on a labeled dataset, which provides the associations or relationships between inputs and outputs. This approach often necessitates a substantial amount of labeled data and significant computational resources for the sake of fine-tuning the parameters of the model.  

Contrarily, in-context learning allows language models to learn tasks based on a few demonstration examples without additional training. This is done by appending demonstration examples to the original prompt following certain criteria \citep{dong2022survey}.  Demonstrations are typically written in natural language which is an interpretable way to communicate with LLMs.

Typically, we organize demonstration examples in a purposeful manner.  A demonstration set consists of multiple examples, each containing a query input text and corresponding label.

In this paper, we omit the structure of the demonstration set.  Instead, we treat both the input prompt and demonstration as mere sets of tokens.  This includes but is not limited to existing ICL prompt structures.  We find this more straightforward mathematically and covers a larger range of scenarios. 
 We consider a standard case where we have an input prompt $X = [X_{1},...,X_{N}]^{T}$ of length $N$ and a Demonstration ICL prompt $X' = [X'_{1},...,X'_{M}]^{T}$ of length $M$. The final input is a concatenation of these two $[X';X]^{T} = [X'_{1},...,X'_{M},X_{1},...,X_{N}]^{T} \in \mathbb{R} ^{(M+N) \times d_{in}}$.  Each $X_{i}$ or $X'_{i}$ represents a single token and they are arranged horizontally as the rows of the matrices.  We refer to the tokens in the ICL prompt $\{ X'_{i} \}_{i=1}^{M}$ as the ICL tokens and the tokens in the input prompt $\{X_{i}\}_{i=1}^{N}$ as the input tokens.

\section{Exact Conversion of In-Context Learning to Model
Weights}
In this section, we formally introduce the ICL weight conversion process. We present that ICL prompts can be captured by a combination of the key and value matrix within a linearized attention setting.

Initially, in section \ref{sec: lin attn naive} and section \ref{section: lin attn free param}, we examine the simplest case with linear attention (Identity Kernel and normalizing factor equalling one).  In section \ref{sec: lin attn naive} we demonstrate that ICL conversion cannot be achieved through modifying existing weights. This is solved in section \ref{section: lin attn free param} through the addition of bias.  Subsequently, in section \ref{sec: assumptions} we outline a list of assumptions.  Transformer models that feature linearized attention and satisfy these assumptions can also achieve bias conversion through the inclusion of bias. Finally, in section \ref{sec: extra param for ext} we formalize the bias terms required for such a linearized attention model.

\subsection{Direct Conversion for Linear Attention} \label{sec: lin attn naive}

In this section, we provide a naive attempt to achieve direct conversion of ICL prompts into the weights of Linear Attention layer.  This involves taking equation \eqref{eqn:lin_attn} and setting $D_{1}(Q_{i})=1, D_{2}(X) = 1$, and $\phi(\cdot) = I(\cdot)$ as the identity mapping.  We want to find corresponding weights $W_{1},W_{2}$ such that the following equivalence holds true for all $X\in \mathbb{R}^{N \times d_{in}}$, 
\begin{equation}
\label{eqn: lin base case}
    W_{1}^{T}X^{T}XW_{2} = W^{T}_{K}X^{T}XW_{V} +W^{T}_{K}X'^{T}X'W_{V}. 
\end{equation}
This is drawn from the vectorized form of Equation \eqref{vectorizedattention}, forming an affine equation that can be rewritten as:
\begin{align}
\label{affinelinearequation}
    &(W^{T}_{2} \otimes W^{T}_{1})\mathrm{vec}(X^{T}X) = \nonumber\\
    &(W^{T}_{V} \otimes W^{T}_{K}) \mathrm{vec}(X^{T}X) + (W^{T}_{V} \otimes W^{T}_{K})\mathrm{vec}(X'^{T}X').
\end{align}
Since $X=0$ is a potential prompt, this means that one would have to have $(W^{T}_{V} \otimes W^{T}_{K})\mathrm{vec}(X'^{T}X') = 0$, which is not true in general. We further explore possible assumptions to achieve such a conversion in the appendix~\ref{app:naive} but ultimately are unable to find a feasible option.  Under very specific circumstances it may be possible to directly convert ICL prompts into the model weights. However, it is likely such circumstances will not apply to real-world scenarios so we will not examine this direction further.

\subsection{Addition of Free Parameters} \label{section: lin attn free param}
Rather than direct conversion, another way to achieve equivalence in equation (\ref{eqn: lin base case}) is through the inclusion of free parameters.  Due to the linearity of the attention module, bias arises as a very natural option. Typically this is included separately within the Key, Value and Query matrices in the form $W_{K} + b_{K}$,$W_{V} + b_{V}$ and $W_{Q} + b_{Q}$ and has limited effect on model performance \citep{namazifar2023role}.

Previously, we mentioned that the Key-Value matrix is integral to the incorporation of ICL prompts.  A natural extension of this principle is to include bias to the Key-Value matrix itself.  We refer to this term as $b_{KV}$.  This makes the linear attention with bias $b_{KV}$:
\begin{equation}
    \mathrm{LinAttn}(Q,K,V,b_{KV}) = Q(K^{T}V + b_{KV}).
\end{equation}
This enables a simple conversion for a given ICL weight as follows.

\begin{theorem}[Bias conversion for Linear Attention]
    For weights $W_{Q},W_{K},W_{V}$, input prompt $X$, ICL prompt $X'$ and initial bias $b_{VK}$, we have that:
    \begin{equation}
    \begin{split}
        \mathrm{LinAttn}([X';X]W_{Q},[X';X]W_{K},[X':X]W_{V},b_{V,K}) \\ = \mathrm{LinAttn}(XW_{Q},XW_{K},XW_{V},b'_{V,K})
    \end{split}
    \end{equation}
    holds for all $X$ if we set
    \begin{equation}
        b'_{V,K} = b_{V,K} + W_{K}^{T}X'^{T}X'W_{V}.
    \end{equation}
\end{theorem}
The proof is trivial by definition.
This conversion is simple because one does not need to consider causal masks and positional encodings. 

\subsection{Assumptions Necessary for Conversion} \label{sec: assumptions}
In this section, we explore what models other than simple linear attention are appropriate for our ICL conversion method.  A similar bias term can be found for any linearized attention model that satisfies all of these requirements.

\begin{assumption}[Autoregressive]
    For any transformer model, we assume that it is autoregressive. At each position, the model can only attend to that token and previous tokens.  This prevents the model from seeing into the future.
\end{assumption}

The autoregressive property of the model is essential because it prevents the ICL tokens from being affected by the input tokens. If the model isn't autoregressive, the ICL tokens themselves will be affected by the input prompt from layer to layer. If the ICL tokens vary depending on the input prompts, then a conversion satisfying one input prompt may not satisfy another.  Most popular LLM models, especially in NLP are already autoregressive by nature so this assumption is easily fulfilled.

\begin{assumption}[Relative Positional Encoding]
 \label{assumption:autoregressive}
    We assume that the transformer model we are dealing with utilizes relative positional encoding.  This means that a token's positional embedding depends not on its absolute position, but rather on its position relative to the other tokens. 
\end{assumption}

This assumption is not an absolute necessity but makes the conversion process significantly easier.  When using absolute positional embedding, there needs to be a linear transformation following the initial input layer to correct for the positional shift. Without such a linear transformation, the inclusion of ICL prompts will change the value of the input prompts and make the conversion more complicated. For simplicity in this study, we will focus on model architectures that include relative positional encodings such as RoFormer \citep{su2023roformer} and Llama \citep{llama}.

\begin{assumption}[Separation property of normalizing factor]
    Assume that for all token sequences A and B, for the given normalizing factor $D_{2}$ in equation(\ref{eqn:lin_attn}) the following is true: $D_{2}([A;B]) - D_{2}([B])=D^{*}_{2}([A])$,  i.e., it only depends on A. 
\end{assumption}

This may seem like a very strong assumption, but it is fulfilled by almost all popular linearized transformer models. Below are a few examples:

\begin{enumerate}
    \item Kernelized attention \citep{katharopoulos2020transformers}:  Set $\phi(\cdot)$ as the representation function of the kernel, set $D_{1}(Q_{i}) = \phi(Q_{i})$ and $D_{2}(X) = \sum_{j=1}^{M}\phi(K'_{i})$. Potential choices for $\phi(\cdot)$ include but are not limited to: \begin{enumerate}
        \item In the original paper \citep{katharopoulos2020transformers} they propose using a simple scheme $\phi(x) = \mathrm{elu}(x) + 1$.
        \item In the Performers architecture by Choromanski et al. \citeyearpar{choromanski2020rethinking}, the PRF (positive random features) and ORF (Orthogonal random features) kernels are learned in the training process.
    \end{enumerate}
    \item RetNet \citep{sun2023retentive}: Set $\phi(\cdot) = I(\cdot)$ be the identity function, set $D_{1}(Q_{i}) = 1$ and $D_{2}(X) = 1$.
\end{enumerate}

\subsection{ICL Conversion for Linearized Attention} \label{sec: extra param for ext}
In this section, we extend the bias conversion method to masked linearized attention layers found in Definition \ref{dfn:masked_lin_attn} with rotary positional encodings (RoPE). Based on Definition \ref{dfn:masked_lin_attn} and the work of Su et al. \citeyearpar{su2023roformer}, kernel methods with RoPE take the form:
\begin{equation}\label{eqn:o_rope}
\begin{split}
    &O^{T}_{i}(X) = \\
    &(R_{\Theta,i}^{d_{K}} \phi(Q_{i}))^{T}\left\{\frac{1}{D_{1}(Q_{i})^{T}D_{2}(X)}
    \sum_{j=1}^{i}R^{d_{K}}_{\Theta,j}\phi(K_{j})V^{T}_{j} \right\},
\end{split}
\end{equation}
where $R_{\Theta,i}^{d_{K}}$ is the rotary matrix defined in the appendix \ref{defn:rotary}.  In the context of positional embeddings, it implies a shift by $i$ positions.

Based on our observations in section \ref{sec:attn}, the logical extension is to include a bias term in the Key-Value matrix. 
 Due to the presence of a normalizing term, we include 2 separate bias terms. 
 The first term $b_{KV}$ is appended to the Key-Value matrix $A=\sum_{j=1}^{i}R^{d_{K}}_{\Theta,j}\phi(K_{j})V^{T}_{j}$.  The second term $b_{D}$ is appended to the normalizing term $D_{2}$.

With the inclusion of bias terms, linearized transformers with RoPE become:
\begin{equation}
\begin{split}
\label{eqn: rope lin attn}
    &O^{T}_{i}(X,b_{KV},b_{D}) =(R_{\Theta,i}^{d_{K}} \phi(Q_{i}))^{T} \\ &\left\{\frac{1}{D_{1}(Q_{i})^{T}(D_{2}(X)+b_{D})}[\sum_{j=1}^{i}R_{\Theta,j}^{d_{K}}\phi(K_{j})V^{T}_{j} + b_{KV}]\right\}.
\end{split}
\end{equation}

\begin{theorem}
\label{main thm}
    Given model weights $W_{V},W_{K},W_{Q}$, input prompt $X$ and ICL prompt $X'$, by taking:
    \begin{equation}
        b'_{KV} = R^{d_{K}}_{\Theta, -M}b_{KV} + \sum_{j=1}^{M}R^{d_{K}}_{\Theta,j-M}\phi(K'_{j})V'^{T}_{j},
    \end{equation} 
    \begin{equation}
        b'_{D} = b_{D} + D^{*}_{2}(X'),
    \end{equation}
    then for all $X$:
    \begin{equation}
        O_{i}([X';X],b_{KV},b_{D}) = O_{i}(X,b'_{KV},b'_{D}).
    \end{equation}
\end{theorem}

The proof is fulfilled in the Appendix \ref{app:proof_thm4.5}.

It is worth highlighting why this makes intuitive sense.  The updated bias $b'_{KV}$ involves first a shift of the original bias term $b_{KV}$ by the rotational matrix $R^{d_{K}}_{\Theta,-M}$, which is the same as a shift left by M tokens. It is equivalent to shifting ``past ICL prompts'' to the left to include new ones when updating the model.

Apart from the practical applications of Theorem \ref{main thm}, we can also arrive at some interesting theoretical conclusions.  Typically, people tend to study the attention-score matrix, which is the interaction between the Query and Key matrices.  However, our study on linearized attention would seem to suggest a profound connection between the Key and Value matrices.  It appears that contextual information is captured within the Key-Value matrix itself.

This yields a whole possible field of research into the interactions between the Key and Value matrix.  Rather than the inclusion of a bias term, perhaps there is merit to other approaches such as scaling terms or using MLPs to approximate the Key-Value matrix itself.  We don't explore this avenue further in this paper, but it is a fascinating direction for future study.

\subsection{ICL Conversion Algorithm for Linearized Attention}
\label{sec:coversion_algorithm}
In this section, we present the algorithm to include ICL prompts $X' = [X'_{1},...,X'_{M}]^{T}$ into a model with linearized attention found in Equation \eqref{eqn: rope lin attn}.  We assume the model has L sub-layers $\{O^{(1)},...,O^{(L)}\}$.  These can include linearized attention, normalization layers, FFN layers, etc.  What's important is except for linearized attention, all the other sub-layers are element-wise operators. 

If $O^{(l)}$ is a linearized attention layer, then we define the weights as $W^{(l)}_{V},W^{(l)}_{K}, W^{(l)}_{Q},b^{(l)}_{KV},b^{(l)}_{D}$.  Algorithm \ref{alg1} takes the form:

\begin{algorithm}[h]
\caption{ICL conversion algorithm  (ICLCA)} \label{alg1}
\begin{algorithmic}[1]
    \STATE \textbf{Input:} $X'^{(0)} = [X'^{(0)}_{1}, ... , X'^{(0)}_{M}]^{T}$
    
    \FOR{$l=1$ {\bf to} $L$}
       \IF{ $O^{(l)}$ is not a linearized attention layer}
            \STATE return $X'^{(l)} = O^{(l)}(X'^{(l-1)})$
        \ELSIF{$O^{(l)}$ is a linearized attention layer with weights $W^{(l)}_{V},W^{(l)}_{K}, W^{(l)}_{Q},b^{(l)}_{KV},b^{(l)}_{D}$}
        \STATE (a) Set $Q'^{(l-1)} = X'^{(l-1)}W^{(l)}_{Q},$ 
      \\ \quad $K'^{(l-1)} = X'^{(l-1)}W^{(l)}_{K}$,   $V'^{(l-1)} = X'^{(l-1)}W^{(l)}_{V}$
        \STATE (b) Set: $b'^{(l)}_{KV} = R^{d_{K}}_{\Theta, -M}b^{(l)}_{KV} +$
      \\ \quad $\sum_{j=1}^{M}R^{d_{K}}_{\Theta,j-M} \phi(K'^{(l-1)}_{j})(V'^{(l-1)}_{j})^{T}$
        \STATE (c) Set: $b'^{(l)}_{D}  = b^{(l)}_{D}+ D^{*}_{2}(X'^{(l-1)})$
        \STATE (d) Set:
      \begin{equation*}
      \begin{split}
          X'^{(l)}_{i}& = \biggl \{(R_{\Theta,i}^{d_{K}}\phi(Q'^{(l-1)}_{i}))^{T} \\&\Bigl [\sum_{j=1}^{i}R_{\Theta,j}^{d_{K}}\phi(K'^{(l-1)}_{j})(V'^{(l-1)}_{j})^{T}+b'^{(l)}_{KV} \Bigl ] \biggl \}/ \\
          & \biggl \{\phi(Q'^{(l-1)}_{i})^{T}\Bigl [ \sum_{j=1}^{i}\phi(K'^{(l-1)}_{j} + b^{(l)}_{D}\Bigl ]\biggl \}
      \end{split}
      \end{equation*}
        \ENDIF
        \ENDFOR
    \STATE Save new bias terms $b^{(l)}_{KV}=b'^{(l)}_{KV}$ and $b^{(l)}_{D} = b'^{(l)}_{D}$ for all $l$.
\end{algorithmic}
\end{algorithm}

\section{Approximate Conversion in Regular Attention} \label{sec: approx}
While our assumptions cover a large number of efficient transformer architectures, a vast body of work is based on the vanilla attention mechanism without linearization found in Equation \eqref{eqn:reg_attn}. We cannot achieve exact conversion of the ICL prompts on such models, but our methods do offer a direct approximation of ICL prompts.

This is done by first approximating the similarity function $\mathrm{sim}(\cdot, \cdot)$ by a kernel and the associated representation function  $\phi(\cdot)$.  The exact conversion is then completed with this approximation using Theorem \ref{main thm} and the bias is reintroduced to the original model.  With the inclusion of RoPE, the resulting output takes the form:
\begin{equation}
    O ^{T}_{i} = \frac{[\sum_{j=1}^{N}\mathrm{sim}(Q_{i},K_{j})V^{T}_{j}] + R_{\Theta,i}^{d_{K}}\phi(Q_{i})^{T}b_{KV}}{[\sum_{j=1}^{N}\mathrm{sim}(Q_{i},K_{j})] + R_{\Theta,i}^{d_{K}}\phi(Q_{i})^{T}b_{D}},
\end{equation}
where
\begin{equation}
    b_{KV} = \sum_{j=1}^{M}R^{d_{K}}_{\Theta,j-M}\phi(K'_{j})V'^{T}_{j}, \quad  b_{D} = \sum_{j=1}^{M}\phi(K'_{j}).
\end{equation}
\setlength{\belowdisplayskip}{3pt}

The accuracy of this approximation depends directly on the accuracy of the kernel when approximating $sim(\cdot, \cdot)$. Even now multiple studies are exploring this subject \citep{choromanski2020rethinking,zhang2024hedgehog} and one can only expect this to get better as time passes.

We include the corresponding algorithm (ICLAA) in Appendix \ref{app:ICLAA}.

\section{Experiments}
We first demonstrate that our ICL conversion algorithm (ICLCA) is numerically exact for linearized attention models. Next we conduct experiments on a synthetic induction head task and show that ICLCA is able to capture the ICL prompt in the bias terms. Furthermore, we perform experiments on a pretrained GPT-2 model and show that, although the conversion process is approximate, our method reduces the relative error of the logits and the model is aware of the context when generating the text. In this section, we denote the model before conversion with ICL prompt by $M_{\mathrm{old, ICL}}$, the model before conversion without ICL prompt by $M_{\mathrm{old, no\ ICL}}$ and the model after conversion without ICL prompt by $M_{\mathrm{new, no\ ICL}}$.

\subsection{Effectiveness of Conversion}
We demonstrate the effectiveness of our method in section \ref{sec: extra param for ext} on a linear attention transformer with RoPE \citep{su2023roformer}. We compute the logits of $M_{\mathrm{old, ICL}}$ and $M_{\mathrm{new, no\ ICL}}$. In Table~\ref{tab:relative_error}, we show the relative error of the output logits before and after the conversion process for different model sizes. It demonstrates that the conversions are indeed exact up to the numerical rounding errors. 
\begin{table}[H]
    \centering
    \begin{small}
    \caption{Relative error of logits. The relative error is computed on the average of 100 randomly generated  ICL prompts and input prompts. 
    }
    \begin{tabular}{c|c|c|c|c|c}
    \toprule
      model size &205K&1.99M&19.8M&198M  & 1.98B \\
      \hline
            relative error   &2.9e-7&4.4e-7&8.3e-7&1.7e-6& 4.3e-6\\
    \bottomrule
    \end{tabular}
    \label{tab:relative_error}
    \end{small}
\end{table}

\subsection{Induction Head Task}
\paragraph{Task}
We show the application of our method on the following induction head task proposed by Bietti et al. \citeyearpar{bietti2023birth} which demonstrates the in-context learning ability. Consider the bigram data model with \textit{trigger tokens} where the sequences follow a mixed Markov distribution. 
To be more specific, the vocabulary is all the letters with both upper and lower cases. Out of the alphabet, the trigger tokens are set to be $\{a, b, c,d,e\}$. 
In each sequence, the tokens form a Markov chain where the transition from non-trigger tokens will be uniformly distributed across the vocabulary while the transition from trigger tokens will be deterministic. 
For instance, in one sequence, when `$a$' first appears, what follows is purely random, but once chosen (say `$x$'), the next time we have `$a$', we must have the same token (`$x$') following(we say that the trigger token `$a$' has committed to token `$x$'). 
This case is closely related to induction \citep{elhage2021mathematical}.
In-context learning on this toy data is very straightforward, i.e., if the context contains committed trigger tokens, then such commitment should be carried over to the rest of the sequence.

\paragraph{Result}
We use a 12-layer linear attention transformer with embedding dimension=128 and RoPE. We evaluate the in-context learning accuracy which is calculated on every committed trigger token. In Table \ref{tab:In_context_accuracy}, we show the in context accuracy under three settings and demonstrate that the ICL prompt is exactly captured by ICLCA. 
\begin{table}[H]
\centering
    \caption{In context accuracy of three models.}
    \begin{tabular}{c|c}
    \toprule
      model & In context accuracy\\
       \hline
      $M_{\mathrm{old, ICL}}$&99.95\%\\
      $M_{\mathrm{old, no\ ICL}}$&2.23\%\\
      $M_{\mathrm{new, no\ ICL}}$&99.95\% \\
      \bottomrule
    \end{tabular}
    \label{tab:In_context_accuracy}
\end{table}
\subsection{Effect of the Added Bias Term}
\begin{figure}[t]
\includegraphics[width=0.48\textwidth]{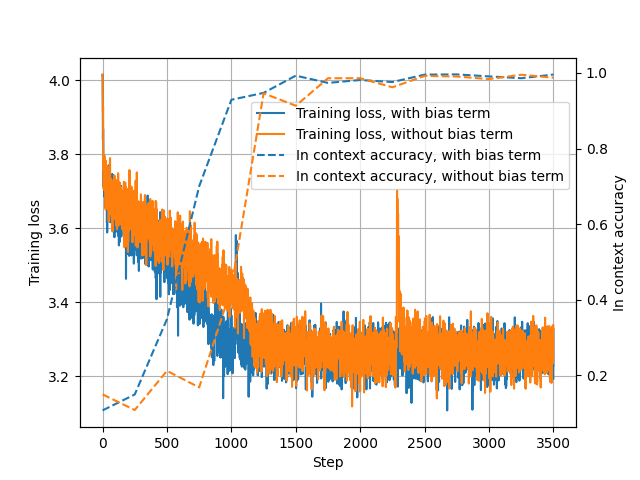}
\caption{The training loss curves and in context accuracy curve during training.}
\label{fig:training}
\end{figure}

In section~\ref{sec: extra param for ext}, we introduce a new model architecture where we add a bias term in the linearized attention module. To show how this bias term affects the model performance, we train the model of the new architecture from scratch.  In Figure~\ref{fig:training} we plot the training loss curves and in-context accuracy curves of models with and without this modification. We see that adding these bias terms makes training slightly faster, which means that our new architecture not only facilitates the ICLCA, but also accomplishes the induction head task better by itself.

\subsection{Experiment on Pretrained GPT-2}
We have shown the effectiveness of ICLCA on the linearized attention models. However, in practice, most existing models are still based on softmax attention. So we also conduct experiments on the pretrained GPT-2 model. Since ICLCA is only valid for the linear attention model, the conversion process contains two steps: 1. First use the method in Performers \cite{choromanski2020rethinking} to approximate the softmax attention by linearized attention; 2. Second use ICLAA (details in Appendix~\ref{app:ICLAA}) to convert the ICL prompt into the bias terms and include these in the original softmax attention. Since the approximation of the softmax attention is not exact, the conversion process introduces some errors. Before the conversion process, the relative error of the logits between $M_{\mathrm{old, ICL}}$ and $M_{\mathrm{old, no\ ICL}}$ is 16.56\%. 
After the conversion process, the relative error of the logits between $M_{\mathrm{old, ICL}}$ and $M_{\mathrm{new, no\ ICL}}$ is 9.17\%.
The relative error of the logits is improved from 16.56\% to 9.17\%. In Table~\ref{tab:ex1} we present an example of the generated text from the three models. This example shows that the conversion method can capture the information in the ICL prompt. We present more examples in Appendix~\ref{app:example}.

\begin{table}[h]
\small{
    \centering
    \caption{Comparison of generated texts.}
    \begin{tabular}{p{7.5cm}}
    \toprule
      \textbf{ICL prompt:} Bill Gates was born at Seattle, Washington. \rule[-3ex]{0pt}{0pt}\\ 
      \textbf{\underline{Model before conversion, with ICL prompt}}\rule[-1.5ex]{0pt}{0pt}\\ 
      \textbf{Input Prompt:} Bill Gates was born at Seattle, Washington. He is the CEO of \rule[-1ex]{0pt}{0pt}\\
      \textbf{Generated Text:} Bill Gates was born at Seattle, Washington. He is the CEO of Microsoft, the world's largest software company. He is also the co-founder and CEO of Microsoft.com, a leading online marketplace for developers and publishers. He is also the co-founder and CEO of Microsoft's Bing search engine. \rule[-2ex]{0pt}{0pt}\\

      \textbf{\underline{Model after conversion, without ICL prompt}} \rule[-1.5ex]{0pt}{0pt}\\ 
      \textbf{Input Prompt:} He is the CEO of \rule[-1ex]{0pt}{0pt}\\
      \textbf{Generated Text:} He is the CEO of Microsoft, the world's largest company. He is a member of the Microsoft Foundation. He is a member of the Council of Economic Advisers.  \rule[-2ex]{0pt}{0pt}\\
      \textbf{\underline{Model before conversion, without ICL prompt}} \rule[-1.5ex]{0pt}{0pt}\\ 
      \textbf{Input Prompt:} He is the CEO of  \rule[-1ex]{0pt}{0pt}\\
      \textbf{Generated Text:} He is the CEO of European Cryptocurrency Exchange, where he rapidly explores the various urban applications for cryptocurrencies and cryptocurrencies in development. The Crypto Central Service crew is looking for cash and other cryptocurrencies based on ethics or finance.\\
      \bottomrule
    \end{tabular}
    \label{tab:ex1}
    }
\end{table}
\section{Transformers as RNNs and Associated Extensions}
Based on the work of Katharopoulos et al. \citeyearpar{katharopoulos2020transformers}, we find that linearized transformers can be expressed in the form of an RNN.  This holds true in our setting where Equation \eqref{eqn:o_rope} becomes:
\begin{equation}
    \begin{split}
    &s_{0} = 0, \quad s_{i} = s_{i-1} + R^{d_{K}}_{\Theta,i}\phi(K_{i})V_{i} \\
    &z_{0} = 0, \quad z_{i} = z_{i-1} + D^{*}_{2}(X_{i}) \\
    &y_{i} = \frac{R^{d_{K}}_{\Theta,i}\phi(Q_{i})^{T}s_{i}}{D_{1}(Q_{i})^{T}z_{i}}.
    \end{split}
\end{equation}

Within this setting, the addition of our bias terms $b_{KV}$ and $b_{D}$ are equivalent to changing the initializations of $s_{0}$ and $z_{0}$ to become: $s_{0} = b'_{KV} - b_{KV} = \sum_{i=j}^{M}R^{d_{K}}_{\Theta,j-M}\phi(K'_{j})V'^{T}_{j}$ and $z_{0} = b'_{D} - b_{D} = D_{2}(X')$.  In RNNs, future tokens can only observe the hidden states of previous tokens, hence capturing the hidden states of the ICL prompts is sufficient.

Viewing ICL conversion from the perspective of RNNs further allows us to generalize to modern state-of-the-art architectures which may not be based on the transformer.  Recently, RNN-based architectures such as H3 \citep{fu2023hungry} and MAMBA \citep{gu2023mamba} have garnered much popularity with many believing they are the next step in LLMs.  While it is yet to be seen whether these models will display emergent abilities such as In-context Learning, if they do, we could extend our ICL conversion method to those autoregressive RNN architectures and capture the ICL prompts in the initialization of the RNN.

\section{Conclusion and Discussion}
In this work, we leverage the linearity of the attention module in linearized attention architectures to permanently incorporate ICL prompts into the model. This is done through the addition of a bias term to the Key-Value matrix, which can be seen as an architectural modification.  Compared to current alternatives, ours is an exact conversion, which is interpretable and very computationally inexpensive to implement.  

Exact conversion from context data to model weights not only contributes to the theoretical understanding of ICL but also unlocks new possibilities for LLMs in practice. 
For instance, LLMs offered by companies are constantly being fine-tuned to incorporate new knowledge, which requires a large amount of data and delicate optimization strategies to avoid catastrophic forgetting. With our method, we can conduct ICL-guided fine-tuning, which can effectively make localized modifications. 
Furthermore, our method can potentially offer significant computation savings for LLM inference, by converting long context prompts such as system prompts, personalized data, and personal history that are repeatedly added as context for every inference. 

When it comes to approximate conversion with standard transformers, being able to add context information in such a cheap way presents new opportunities.  While the approximate conversion itself may not be sufficient for real-world applications, it can be used as a potential initialization for existing fine-tuning methods to reduce the number of training steps required.  

A legitimate concern is the scalability of our method with model size. In the case of linear attention models, the conversion is exact so the methodology scales without any issues to large models. In Table \ref{tab:relative_error}, we show that rounding errors scale well with the number of model parameters. As the model size increases by a factor of 10, the rounding errors increase by approximately a factor of 2. Regarding the approximate softmax conversion method, scalability is an issue. That being said, the scalability depends primarily on whether the kernel function can properly approximate the softmax in pre-trained transformers. If such kernels can scale to larger models our approximate conversion methodology will scale to the same models using that kernel. In fact, in the original works the approximation is done for all tokens. In this work we only approximate softmax for the ICL tokens. Hence the precision of our method should always be better than approximating the full softmax with all the tokens. Recent work such as Hedgehog and Porcupine \citep{zhang2024hedgehog} has shown it is possible to design kernels that scale fairly well to large models such as LlaMa-2. Furthermore, we think it is possible to design kernels specifically for ICL approximation which will perform even better for this specific task. We believe with the correct choice of kernel, our conversion method can scale to large models even for the approximate case.

A natural extension to these findings is to test our approximate conversion method with more sophisticated kernels \cite{katharopoulos2020transformers,zhang2024hedgehog}.  Further experiments can also be conducted to verify how the approximate conversion works in conjunction with existing fine-tuning methods.  Another possible line of research is exploring possible methods to achieve exact conversion for transformers with softmax attention.

\section*{Impact Statement}
This paper presents work whose goal is to advance the field of large language models. By enabling permanent and interpretable incorporation of context into large language models without expensive parameter updates, this approach can make advanced AI more accessible and cost-effective. This democratization can enhance various applications, including education, customer service, and content creation. However, it also raises concerns about potential misuse, such as the reinforcement of biases or misinformation. Therefore, it is crucial to establish ethical guidelines and safeguards to ensure that the benefits of this technology are realized responsibly and equitably.

\section*{Acknowledgement}
We would like to thank Shuchen Xue, Tianxun Zhou,  Tianyi Zhang, Liu Wei, for the helpful discussions and suggestions.

This research is partially supported by the National Research Foundation Singapore under the AI Singapore Programme (AISG Award No: AISG2-TC-2023-010-SGIL) and the Singapore Ministry of Education Academic Research Fund Tier 1 (Award No: T1 251RES2207).

\bibliography{reference}
\bibliographystyle{icml2024}

\newpage
\appendix 
\onecolumn
\section*{\Large Appendix}
\section{Naive Attempt at Direct Conversion} \label{app:naive}
To construct the equivalence in equation (\ref{eqn: lin base case}) additional assumptions and properties have to be included. Firstly, the term $X^{T}X$ does not include the entire matrix space.  All matrices of the form $X^{T}X$ are positive semi-definite.  Hence, we can limit the matrix space to only matrices of the form $U^{T}\Sigma U$ where $U^{T},U$ are orthogonal matrices and $\Sigma$ is a diagonal matrix where each term on the diagonal is the eigenvalue.

Such a space still contains the zero vector.  Equation (\ref{affinelinearequation}) shows that the equivalence does not accommodate for the zero vector.  Another possible assumption is for the token set to belong to a shifted subspace.  i.e. For all inputs $X$, we have:
\begin{equation}
    X^{T}X \in c + D.
\end{equation}

Here $c$ is a shift vector that satisfies equation (\ref{affinelinearequation}), i.e. 
\begin{equation}
    (W^{T}_{1} \otimes W^{T}_{2})vec(c) = (W^{T}_{V} \otimes W^{T}_{K}) vec(c) + (W^{T}_{V} \otimes W^{T}_{K})vec(X'^{T}X').
\end{equation}

At the same time we need $D \subset ker(W^{T}_{1}\otimes W^{T}_{2} - W^{T}_{V}\otimes W^{T}_{K}).$

There are 2 key problems with such a setting.  First of all, if $W_{1},W_{2}$ exists to satisfy such a scenario, there is no guarantee for the uniqueness.  In fact, it is likely not to be unique and there is no clear intuitive choice.  Furthermore, $c$ depends not on $X'$ but rather on $X$.  That would mean that the matrix space assumption would vary depending on $X$. This is very difficult because the equality must be true for all X.  Hence we choose to consider the addition of free parameters rather than attempt a direct solution.
\section{Proof of Theorem \ref{main thm}}
\label{app:proof_thm4.5}
\begin{proof}
From the original paper introducing RoPE \citep{su2023roformer}, the rotational matrix is defined as:

\begin{definition}[Rotary matrix] \label{defn:rotary}
    A rotary matrix with pre-defined parameters $\Theta = \{\theta_{i} = 10000^{-2(i-1)/d},i\in[1,...,d/2]\}$ and shift m is defined as:
    \begin{equation}
    R^{d}_{\Theta,m} = 
\begin{pmatrix}
\cos m\theta_1 & -\sin m\theta_1 & 0 & \cdots & 0 & 0 & 0\\
\sin m\theta_1 & \cos m\theta_1 & 0 & \cdots & 0 & 0 & 0 \\
0 & 0 & \cos m\theta_2 & -\sin m\theta_2 & \cdots & 0 & 0 \\
0 & 0 & \sin m\theta_2 & \cos m\theta_2 & \cdots & 0 & 0 \\
\vdots & \vdots & \vdots & \vdots & \ddots & \vdots & \vdots \\
0 & 0 & 0 & 0 & \cdots & \cos m\theta_{d/2} & -\sin m\theta_{d/2} \\
0 & 0 & 0 & 0 & \cdots & \sin m\theta_{d/2} & \cos m\theta_{d/2}.
\end{pmatrix}
\end{equation}
\end{definition}

As one would expect with rotational matrices, we have:

\begin{equation}
    R^{d}_{\Theta,a+b} = R^{d}_{\Theta,a} + R^{d}_{\Theta,b} ,
\end{equation}

and 
\begin{equation}
    R^{d \: T}_{\Theta , m} = R^{d}_{\Theta, -m}.
\end{equation}

\begin{equation}
    \begin{split} 
    O^{T}_{i}([X',X],b_{1},b_{2}) &= \frac{(R_{\Theta,i+M}^{d_{K}}\phi(Q_{i}))^{T}[\sum_{j=1}^{i}(R_{\Theta,j+M}^{d_{K}}\phi(K_{j}))V^{T}_{j}+ \sum_{j=1}^{M}(R_{\Theta, j}^{d_{K}}\phi(K'_{j}))V'^{T}_{j} + b_{1}]}{D_{i}([X';X])+b_{2}} \\
    &= \frac{\phi(Q_{i})^{T}[\sum_{j=1}^{i}R_{\Theta,i}^{d_{K} \; T}(R_{\Theta,j}^{d_{K}}\phi(K_{j})V^{T}_{j} + \sum_{j=1}^{M}R_{\Theta,i}^{d_{K} \; T}(R_{\Theta,j-M}^{d_{K}}\phi(K'_{j}))V'^{T}_{j} +R_{\Theta,i}^{d_{K} \; T}R_{\Theta,-M}^{d_{K}}b_{1}]}{D_{i}(X) + D^{*}_{i}(X')+b_{2}} \\
    & = \frac{(R_{\Theta,i}^{d_{K}}\phi(Q_{i}))^{T}[\sum_{j=1}^{i}(R_{\Theta,j}^{d_{K}}\phi(K_{j})V^{T}_{j} + \sum_{j=1}^{M}(R_{\Theta,j-M}^{d_{K}}\phi(K'_{j}))V'^{T}_{j} +R_{\Theta,-M}^{d_{K}}b_{1}]}{D_{i}(X) + b'_{2}} \\
    &= O^{T}_{i}(X,b'_{1},b'_{2}).
    \end{split}
\end{equation}
\end{proof}
\section{Analysis of Other Choices of Free Parameters }

In this section, we consider other options for free parameters within a standard transformer with stacked decoders.  Free parameters can be included in 3 main sections.  The input, attention module, and feed-forward network.  To begin with, we can modify the input to the Attention Module.  Since GPT networks and many other LLM architectures are stacked layers of decoders, all of the information included by ICL tokens could technically be captured in an additional encoding layer before the input and transformer layers.  There are interesting approaches we considered, such as a modification of the current compressor architecture in popular models such as transformers-XL\citep{dai2019transformer}, effectively compressing the ICL inputs and using it as an initialization.  However, these seem like fanciful ways to append the ICL prompts to the start of all future inputs

An alternative is to directly modify the FFN.  This could potentially be very interesting because past work on ROME \citep{meng2022locating}  has demonstrated that factual associations within GPT networks can be located in the MLP/FFN module of the attention process.  Instead of adding bias terms to the numerator and denominator in theorem \ref{main thm}, we could attempt to induce a similar change within the FFN.  While this would be architecturally clean, an exact conversion could still be a challenge.

In this study, we opted to focus on the addition of free parameters to the attention module.  Apart from a bias term, there were a few other methods considered.  First of all, is a scaling term.  However, since the input tokens can be similarly scaled, that shouldn't help with conversion.  Another attempt was to include starting/initialization tokens within the key value matrix, however this would just give the same result as the bias term but with more computational cost.

A possible option to look at in future studies is to include the full FFN within the Key-Value matrix itself.  This marks a clear departure from the current transformer architecture but would be consistent with our finding that the Key-Value matrix bears special significance.  However, such claims will require detailed experiments for verification.

\section{Extension to Other Architectures}
In this paper, we choose to consider different modifications of the typical stacked transformer architecture.  A future alternative to this would be GPT-J \citep{gpt-j}, which places the attention module and feed-forward network in tandem rather than one after the other.

GPT-J hasn't seen widespread implementation because, while it matches the typical stacked architecture in performance, it fails to exceed it.  However, it provides a very natural structure for ICL conversion.  Rather than include a whole new bias term to the attention module, one can add that term to the FFN itself.  i.e. for each token $\textbf{q}$, rather than just taking $FFN(\textbf{q})$, we would take $FFN(\textbf{q}) + \textbf{q}\sum_{j=1}^{M} R^{d_{K}}_{\Theta,j-M}\phi(K'_{j})^{T}V'^{T}_{j}$.  

There are a few problems to address before such an approach becomes viable.  First of all, GPT-J uses the traditional non-linearized attention with softmax.  There has yet to be a detailed study into how it would behave in a linearized setting which is a prerequisite to our conversion method.  More importantly, any linear setting introduced would have to have a normalizing term $D(\cdot) = 1$ like in RetNet.  While RetNet has shown great promise so far, it may not necessarily be the future of transformer networks.  Hence being limited to cases with $D(\cdot) = 1$ is undesirable.

\subsection{ICL Approximation Algorithm for Regular Attention} \label{app:ICLAA}
In this section we present the algorithm to approximately include ICL prompts $X' = [X'_{1},...,X'_{M}]^{T}$ into a model with regular attention found in equation (\ref{eqn:regular attention}).  We assume the model has L sub-layers $\{O^{(1)},...,O^{(L)}\}$.  These can include regular attention, normalization layers, FFN layers, etc.  Except for regular attention, all other sub-layers are element-wise operators. 

If $O^{(l)}$ is a regular attention layer, then we define its weights as $W^{(l)}_{V},W^{(l)}_{K}, W^{(l)}_{Q}$.  Algorithm \ref{alg2} takes the form:

\begin{algorithm}[H]
\caption{ICL approximation algorithm  (ICLAA)}\label{alg2}
\begin{algorithmic}[1]
  \STATE \textbf{Input:} $X^{\prime(0)} = [X^{\prime(0)}_{1}, ... , X^{\prime(0)}_{M}]^{T}$.
  \FOR{$l=1$ {\bfseries to} $L$}
 \IF{ $O^{(l)}$ is not a regular attention layer} 
  \STATE return $X^{\prime(l)} = O^{(l)}(X^{\prime(l-1)})$
  \ELSIF{$O^{(l)}$ is a regular attention layer with weights $W^{(l)}_{V},W^{(l)}_{K}, W^{(l)}_{Q}$}
  \STATE (a) Set $Q^{\prime(l-1)} = X^{\prime(l-1)}W^{(l)}_{Q}$, $K^{\prime(l-1)} = X^{\prime(l-1)}W^{(l)}_{K}$,   $V^{\prime(l-1)} = X^{\prime(l-1)}W^{(l)}_{V}$;
  \STATE (b) Set $b^{(l)}_{KV} = \sum_{j=1}^{M}R^{d_{K}}_{\Theta,j-M}\phi(K^{\prime(l-1)}_{j})(V^{\prime(l-1)}_{j})^{T}$;
  \STATE (c) Set $b^{(l)}_{D}  = \sum_{j=1}^{M}R^{d_{K}}_{\Theta,j-M}\phi(K^{\prime(l-1)}_{j})$;
  \STATE (d) Set $x^{\prime(l)}_{i} =\biggl \{\sum_{j=1}^{i}\mathrm{sim}(R^{d_{K}}_{\Theta,i}Q^{\prime(l-1)}_{i},R^{d_{K}}_{\Theta,j}K^{\prime(l-1)}_{j})V^{\prime(l-1)
  \; T}_{j}\biggl \}/\biggl \{ \sum_{j=1}^{i}\mathrm{sim}(Q^{\prime(l-1)}_{i},K^{\prime(l-1)}_{j})\biggl \}$.
  \ENDIF
  \ENDFOR
  \STATE Save new bias terms $b^{(l)}_{KV}$ and $b^{(l)}_{D}$ for all $l$.
\end{algorithmic}
\end{algorithm}

\section{More Examples of GPT-2 Generated Text}
\label{app:example}
In this section, we present more examples of GPT-2 generated text before and after conversion.
\begin{table}[H]
\small{
    \centering
    \caption{Comparison of generated texts before and after conversion}
    \begin{tabular}{p{15cm}}
    \toprule
      \textbf{ICL prompt:} Los Angeles is the second most populous city in the United States. \rule[-3ex]{0pt}{0pt}\\ 
      \textbf{\underline{Model before conversion, with ICL prompt}}\rule[-1.5ex]{0pt}{0pt}\\ 
      \textbf{Input Prompt:} Los Angeles is the second-most populous city in the United States. It is located in the U.S. state of \rule[-1ex]{0pt}{0pt}\\
      \textbf{Generated Text:} Los Angeles is the second-most populous city in the United States. It is located in the U.S. state of California, which is home to the largest number of residents of any U.S. state. \rule[-2ex]{0pt}{0pt}\\

      \textbf{\underline{Model after conversion, without ICL prompt}} \rule[-1.5ex]{0pt}{0pt}\\ 
      \textbf{Input Prompt:} It is located in the U.S. state of\rule[-1ex]{0pt}{0pt}\\
      \textbf{Generated Text:} It is located in the U.S. state of California. The city of Los Angeles is the second most populous city in the United States, with a population of 1.5 million people.  \rule[-2ex]{0pt}{0pt}\\
      \textbf{\underline{Model before conversion, without ICL prompt}} \rule[-1.5ex]{0pt}{0pt}\\ 
      \textbf{Input Prompt:} It is located in the U.S. state of \rule[-1ex]{0pt}{0pt}\\
      \textbf{Generated Text:} It is located in the U.S. state of Texas. The State of Texas has a population of about 3,000,000.
\\
      \bottomrule
    \end{tabular}
    \label{tab:generation2}
    }
\end{table}

\begin{table}[H]
\small{
    \centering
    \caption{Comparison of generated texts before and after conversion}
    \begin{tabular}{p{15cm}}
    \toprule
      \textbf{ICL prompt:} Steve Jobs founded Apple. \rule[-3ex]{0pt}{0pt}\\ 
      \textbf{\underline{Model before conversion, with ICL prompt}}\rule[-1.5ex]{0pt}{0pt}\\ 
      \textbf{Input Prompt:} Steve Jobs founded Apple. He is the founder of \rule[-1ex]{0pt}{0pt}\\
      \textbf{Generated Text:} Steve Jobs founded Apple. He is the founder of Apple, Inc., a leading technology company that has been around for over 40 years. He is also the founder of the Apple logo. \rule[-2ex]{0pt}{0pt}\\

      \textbf{\underline{Model after conversion, without ICL prompt}} \rule[-1.5ex]{0pt}{0pt}\\ 
      \textbf{Input Prompt:} He is the founder of\rule[-1ex]{0pt}{0pt}\\
      \textbf{Generated Text:} He is the founder of the company that created the iPhone. Apple has been a major player in the tech industry for over a decade. Jobs was a pioneer in the Apple Watch. Jobs was a former CEO of Apple.\rule[-2ex]{0pt}{0pt}\\
      \textbf{\underline{Model before conversion, without ICL prompt}} \rule[-1.5ex]{0pt}{0pt}\\ 
      \textbf{Input Prompt:} He is the founder of \rule[-1ex]{0pt}{0pt}\\
      \textbf{Generated Text:} He is the founder of the New York-based group, which has been a vocal critic of the Obama administration's handling of the Syrian conflict. "This is not a war of words," he said. "This is a war of words."

\\
      \bottomrule
    \end{tabular}
    \label{tab:generation3}
    }
\end{table}
\end{document}